\documentclass{article}
\usepackage{tabularx}
\usepackage{microtype}
\usepackage{tcolorbox}
\usepackage{listings}
\usepackage{booktabs}
\usepackage{float}
\usepackage{subcaption}
\usepackage{adjustbox}
\usepackage{graphicx} 
\usepackage{multirow}
\usepackage{hyperref}

\usepackage[preprint]{icml2026}

\usepackage{amsmath}
\usepackage{amssymb}
\usepackage{mathtools}
\usepackage{amsthm}

\usepackage[capitalize,noabbrev]{cleveref}

\theoremstyle{plain}

\theoremstyle{definition}

\theoremstyle{remark}

\usepackage[textsize=tiny]{todonotes}

\icmltitlerunning{DyTopo: Dynamic Topology Routing for Multi-Agent Reasoning via Semantic Matching}

\begin{document}

\twocolumn[
  \icmltitle{DyTopo: Dynamic Topology Routing for Multi-Agent Reasoning via Semantic Matching}



  \icmlsetsymbol{equal}{*}

  \begin{icmlauthorlist}
    \icmlauthor{Yuxing Lu}{equal,pek,gt}
    \icmlauthor{Yucheng Hu}{equal,seu}
    \icmlauthor{Xukai Zhao}{thu}
    \icmlauthor{Jiuxin Cao}{seu}
  \end{icmlauthorlist}

  \icmlaffiliation{pek}{Peking University, Beijing, China}
  \icmlaffiliation{gt}{Georgia Institute of Technology, Atlanta, United States}
  \icmlaffiliation{seu}{Southeast University, Location, Country}
  \icmlaffiliation{thu}{Tsinghua University}

  \icmlcorrespondingauthor{Jiuxin Cao}{jx.cao@seu.edu.cn}

  \icmlkeywords{Machine Learning, ICML}

  \vskip 0.3in
]



\printAffiliationsAndNotice{}  

\begin{abstract}
Multi-agent systems built from prompted large language models can improve multi-round reasoning, yet most existing pipelines rely on fixed, trajectory-wide communication patterns that are poorly matched to the stage-dependent needs of iterative problem solving. We introduce DyTopo, a manager-guided multi-agent framework that reconstructs a sparse directed communication graph at each round. Conditioned on the manager's round goal, each agent outputs lightweight natural-language \emph{query} (need) and \emph{key} (offer) descriptors; DyTopo embeds these descriptors and performs semantic matching, routing private messages only along the induced edges. Across code generation and mathematical reasoning benchmarks and four LLM backbones, DyTopo consistently outperforms over the strongest baseline (avg. +6.2). Beyond accuracy, DyTopo yields an interpretable coordination trace via the evolving graphs, enabling qualitative inspection of how communication pathways reconfigure across rounds.
\end{abstract}

\section{Introduction}
Multi-agent systems built from prompted large language models have become a practical paradigm for multi-round reasoning~\citep{tran2025multi, li2024survey}. By instantiating several role-specialized LLM agents and allowing them to interact over multiple rounds, these systems can iteratively refine partial solutions, cross-check intermediate steps, and integrate complementary skills~\citep{wolflein2025llm, yu2025survey}. This collaboration style is particularly well suited to complex problem domains, where effective reasoning emerges from the coordinated interplay of specialized agents and from their abilities to collectively revise earlier assumptions as new evidence or errors are uncovered.

A central yet often under-specified aspect of multi-agent reasoning is the communication structure: which agents exchange information with whom, and when~\citep{goldman2003optimizing}. Many existing pipelines default to a fixed, trajectory-wide interaction pattern (e.g., broadcast discussion or scripted turn-taking), effectively reusing the same topology across all rounds. However, multi-round reasoning is stage-dependent: early rounds tend to benefit from broad exploration and shared problem framing, whereas later rounds require selective, high-precision exchanges to diagnose failures and converge on a coherent solution~\citep{liu2024autonomous}. This suggests that communication topology should be an adaptive object, conditioned on the round-level goal, rather than a static design choice.

\begin{figure}[t]
    \centering
    \includegraphics[width=\linewidth]{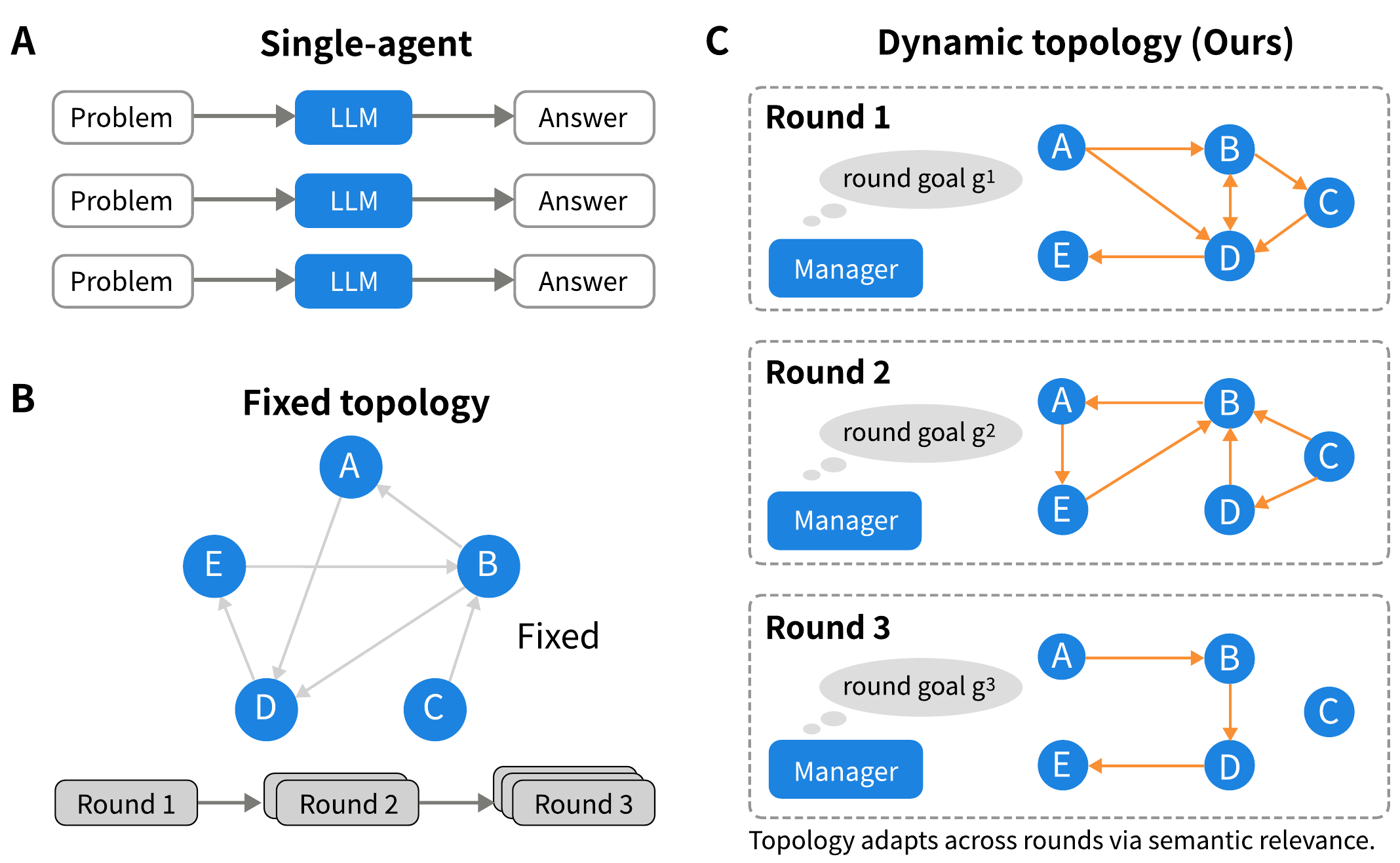}
    \caption{\textbf{Comparison of communication topologies.} (A) Single-agent prompting. (B) Fixed-topology communication reused across rounds. (C) \textbf{DyTopo} dynamically rewires a directed agent graph each round based on the round goal and semantic relevance.}
    \label{fig:teaser}
\end{figure}

We propose \textbf{DyTopo}, a manager-guided multi-round multi-agent framework in which the manager specifies a round-level goal and determines whether to terminate the interaction. Given the round goal and the agents' self-described information needs and capabilities, DyTopo induces a directed communication graph at each round and routes messages only along the activated links. This allows the system to shift from broad exploration to targeted verification as reasoning progresses.

A key ingredient enabling DyTopo’s round-by-round routing is a semantic key-query matching scheme across agents. Conditioned on the manager’s round goal, each agent provides short natural-language descriptors that summarize what it can provide to others (a “key”) and what it currently seeks (a “query”). DyTopo semantically matches queries to keys to induce the directed communication graph for each round, and routes messages only along the activated links. This decouples what agents generate from how their information is routed, enabling communication patterns that adapt over rounds. This topology-driven routing yields two advantages. First, it improves collaboration by organizing information flow around the current round goal rather than a static neighborhood. Second, it provides an interpretable coordination trace: edges are activated based on explicit descriptors and semantic relevance, so the evolving graphs can be inspected to reveal how pathways reconfigure over time and which patterns correlate with success or failure.

We evaluate DyTopo on multi-round code generation and mathematical reasoning tasks, comparing against single-agent prompting, multi-agent baselines with fixed or random communication topologies, and strong recent agentic frameworks. The results show that dynamic communication topologies consistently improve task performance and remain robust under different experiment settings. We further characterize the method through analyses of performance versus the number of rounds, qualitative visualizations of topology evolution over time, and ablations over the semantic matching hyperparameters that control link activation.

\section{Related Work}

\subsection{LLM-Based Multi-Agent Collaboration}
A growing line of work studies how to compose multiple prompted LLM instances into a cooperative system via natural-language interaction. Early frameworks emphasize role specialization and structured dialogue: CAMEL proposes role-playing agents guided by inception prompting to autonomously collaborate on tasks~\citep{li2023camel}, while AutoGen provides a programmable framework for building applications from multiple conversable agents with customizable interaction patterns~\citep{wu2023autogen}. MetaGPT further incorporates human-inspired standardized operating procedures (SOPs) to coordinate multiple role agents and reduce cascading errors in long workflows~\citep{hong2023metagpt}. Complementary to role-based cooperation, multi-agent deliberation improves reasoning and factuality by having multiple model instances propose and critique solutions over multiple rounds~\citep{du2023improving}. Finally, agent systems are often coupled with tools or external models, where an LLM acts as a controller that decomposes tasks and delegates to specialized executors \citep{shen2023hugginggpt}. While these approaches demonstrate gains from collaboration, they typically rely on fixed or dense communication patterns, leaving open how to adaptively route information among agents at inference time.

\subsection{Selective and Dynamic Communication Topologies}
Selective communication has long been studied in multi-agent learning and neural routing. In cooperative MARL, targeted messaging methods such as TarMAC learn \emph{what} to communicate and \emph{whom} to address, enabling multi-round coordination with interpretable communication patterns~\citep{das2019tarmac}. In large-scale neural architectures, conditional computation and routing activate only a small subset of experts per token to scale capacity efficiently~\citep{fedus2021switch}, and content-based sparse attention constructs query-dependent sparse interaction patterns among tokens~\citep{roy2021efficient}. Recently, these principles have been adapted to LLM-based agent teams to reduce redundant interactions and design task-aware connectivity. AgentPrune identifies communication redundancy in multi-agent pipelines and prunes low-value messages on the induced spatio-temporal message-passing graph~\citep{zhang2024agentprune}. Beyond pruning, G-Designer generates task-conditioned agent communication topologies~\citep{zhang2024gdesigner}, and GTD casts topology synthesis as a guided diffusion process to optimize performance-cost-robustness trade-offs~\citep{jiang2025gtd}. Our work complements this direction by studying an explicitly interpretable inference-time routing mechanism: agents output textual Need and Offer descriptors and a directed topology is constructed each round via semantic similarity, enabling controlled multi-round message passing and topology-level analysis.

\section{Methods}
\label{sec:methods}
\begin{figure*}
    \centering
    \includegraphics[width=0.9\textwidth]{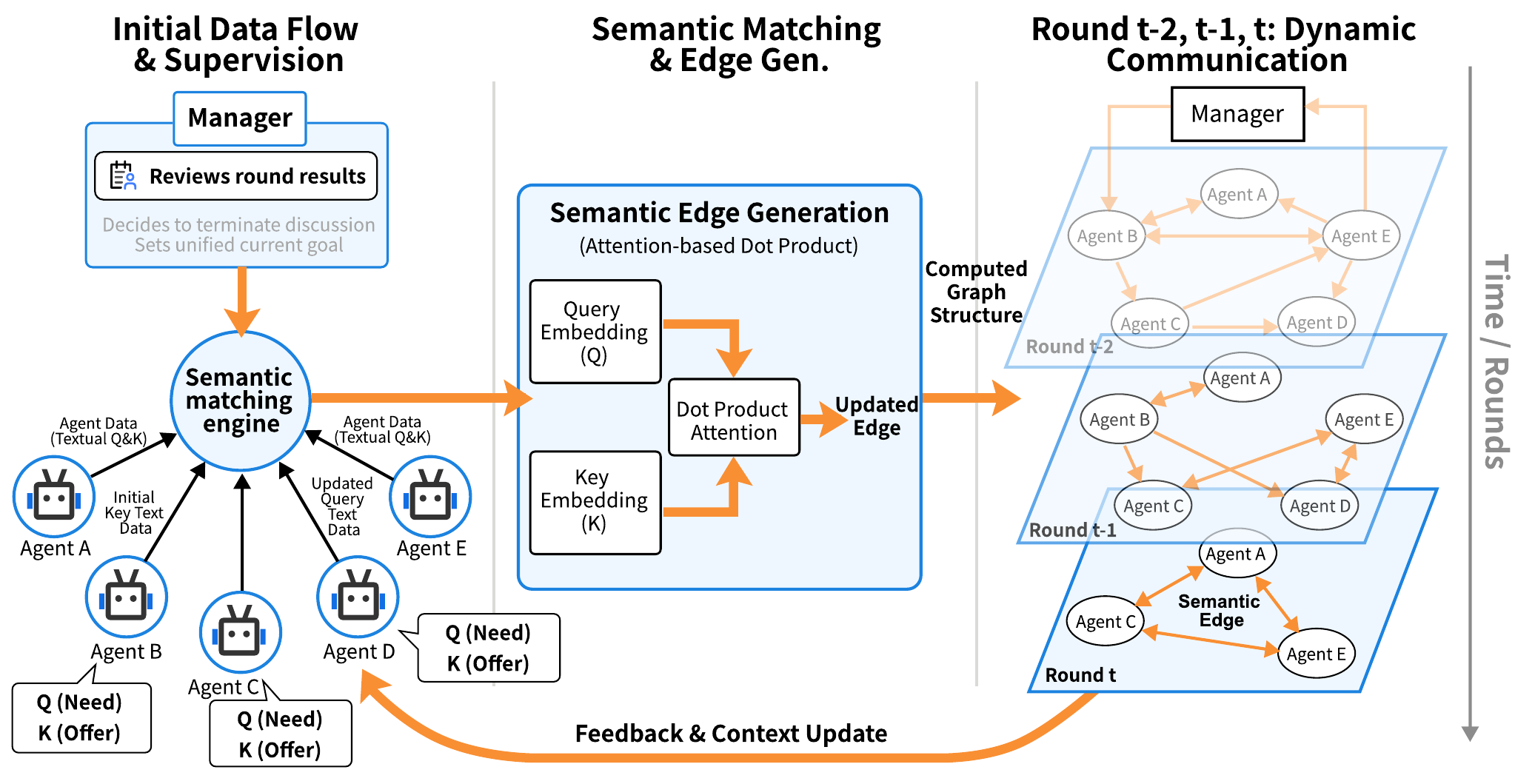}
    \caption{\textbf{DyTopo round-by-round routing via semantic matching.} At each round $t$, each worker agent outputs a query and a key descriptor. A semantic matching module embeds these descriptors, computes pairwise similarity, and induces a directed graph $G^{(t)}$. Private messages produced at round $t$ are routed according to $G^{(t)}$ after a synchronization barrier and are appended to recipients' memories for round $t{+}1$. The Manager provides round goals and updates the next-round context, yielding a closed-loop adaptation across rounds.}
    \label{fig:framework}
\end{figure*}

We formalize DyTopo as a \textbf{Dynamic Computation Graph} (DCG), $\mathcal{G}=\{G^{(t)}\}_{t=0}^{T-1}$, where $T$ is the number of executed rounds (indexed by $t\in\{0,\dots,T-1\}$) and $T\le T_{\max}$ is capped by a fixed budget. Unlike static topologies, DyTopo reconstructs $G^{(t)}$ at each communication round $t$ under a manager-specified round goal, driven by semantic matching between agents' information needs and offered capabilities. In Appendix~\ref{app:algorithm}, we analyze Dytopo's complexity advantages over fully connected networks. The overall algorithm is summarized in Appendix~\ref{sec:Pseudo-code}.

\subsection{Preliminaries}
Let $\mathcal{A}=\{a_1,\dots,a_N\}$ denote $N$ heterogeneous worker agents. Each worker agent $a_i$ is instantiated with a role description $\rho_i$ and maintains a local memory buffer $\mathcal{H}_i^{(t)}$. At each round $t$, agent $a_i$ produces communication messages in $\mathcal{M}$ and lightweight topology descriptors in $\mathcal{D}$. In addition, DyTopo includes a Manager meta-agent that maintains a global view and updates the round context $C_{task}^{(t)}$.

DyTopo uses two message channels: a manager-visible public channel $\mathcal{M}_{pub}$ and a routed private channel $\mathcal{M}_{priv}$. Agent $a_i$ outputs a public message $m^{(t)}_{pub,i}\in\mathcal{M}_{pub}$ (visible to the Manager and recorded for analysis) and a private message $m^{(t)}_{priv,i}\in\mathcal{M}_{priv}$ (routed to the out-neighbors of $a_i$ in $G^{(t)}$).

In addition, the agent outputs two short natural-language descriptors that determine connectivity at round $t$. The query descriptor $s^{(t)}_{q,i}\in\mathcal{D}$ summarizes what information agent $a_i$ currently seeks, and the key descriptor $s^{(t)}_{k,i}\in\mathcal{D}$ summarizes what information it can provide to others.

\subsection{Per-Round Agent Execution}
\subsubsection{Single-Pass Inference and Descriptor Generation}
To ensure computational efficiency, we impose a Single-Pass Inference constraint: each agent performs exactly one forward pass per round using only its role description, the manager-provided round goal, and its local memory. Agents generate task-relevant messages and lightweight natural-language descriptors used for topology induction.

Formally, the local state is
\begin{equation}
\label{eq:state_definition}
S_i^{(t)} = [\rho_i;\, C_{task}^{(t)};\, \mathcal{H}_i^{(t)}],
\end{equation}
and the agent output is
\begin{equation}
O_i^{(t)}=\langle m^{(t)}_{pub,i},\, m^{(t)}_{priv,i},\, s^{(t)}_{q,i},\, s^{(t)}_{k,i}\rangle
\sim \pi_{\theta_i}(\cdot \mid S_i^{(t)}).
\end{equation}
The descriptors $s^{(t)}_{q,i}$ and $s^{(t)}_{k,i}$ are embedded to induce $G^{(t)}$ (Sec.~\ref{sec:dynamic_topology}); private messages are then routed and integrated into memories for the next round (Sec.~\ref{sec:state_update}).

\subsubsection{Synchronization Barrier and Context Update}
\label{sec:state_update}
After generating $O_i^{(t)}$, agents do not update their local memory immediately. Instead, DyTopo applies a Synchronization Barrier: it first induces the directed topology $G^{(t)}$ and routes private messages according to the activated edges, and only then updates each agent's memory for next round.

Let $G^{(t)}=(\mathcal{A},\mathcal{E}^{(t)})$ and define the incoming neighbor set as $\mathcal{N}^{(t)}_{in}(i)=\{\,j \mid (a_j \rightarrow a_i)\in \mathcal{E}^{(t)}\,\}$. 

The memory update rule is:
\begin{equation}
\small
\mathcal{H}_i^{(t+1)}=
\mathcal{H}_i^{(t)}
\oplus m_{pub,i}^{(t)}
\oplus \Sigma_{\sigma_i^{(t)}}\!\left(\{m_{priv,j}^{(t)} \mid j\in \mathcal{N}^{(t)}_{in}(i)\}\right).
\end{equation}
where $\oplus$ denotes concatenation and $\mathcal{N}^{(t)}_{in}(i)$ is the set of incoming neighbors of agent $a_i$ in $G^{(t)}$. $\Sigma_{\sigma^{(t)}}(\cdot)$ is a context aggregation operator that constructs a single prompt block from routed private messages by ordering them according to an \emph{aggregation order} $\sigma^{(t)}$ and then concatenating them in that order. This yields a deterministic prompt layout and an ordering that is consistent with the induced dependency structure. Here, $m_{pub,i}^{(t)}$ is agent $a_i$'s own generated public message at round $t$, and each $m_{priv,j}^{(t)}$ is included in agent $a_i$'s next-round context only when $j$ is an incoming neighbor (i.e., when the edge $a_j\!\rightarrow\! a_i$ is active in $G^{(t)}$). Overall, this update ensures that agent $a_i$ at round $t{+}1$ conditions only on its prior memory and on information permitted by the induced topology $G^{(t)}$.

\subsection{Dynamic Topology via Semantic Matching}
\label{sec:dynamic_topology}

DyTopo induces a directed communication graph at each round based on the agents' textual descriptors. Concretely, at round $t$ each agent $a_i$ outputs a query descriptor $s^{(t)}_{q,i}$ (what it needs) and a key descriptor $s^{(t)}_{k,i}$ (what it can provide). DyTopo embeds these descriptors into a shared semantic space and constructs $G^{(t)}=(\mathcal{A},\mathcal{E}^{(t)})$ by activating directed edges from providers to consumers according to semantic relevance. The resulting topology determines which private messages are routed between agents in round $t$.

\subsubsection{Semantic Alignment Quantification}
\label{sec:semantic_alignment}

Because descriptors are natural language, we map them to vectors using a fixed pre-trained semantic encoder,
$\mathrm{Emb}:\mathcal{D}\rightarrow\mathbb{R}^d$, where $d$ is the embedding dimension.
For each agent $a_i$ at round $t$, we compute
\begin{equation}
\mathbf{q}_i^{(t)}=\mathrm{Emb}\!\left(s^{(t)}_{q,i}\right),\quad
\mathbf{k}_i^{(t)}=\mathrm{Emb}\!\left(s^{(t)}_{k,i}\right),
\end{equation}
and stack them into matrices $\mathbf{Q}^{(t)},\mathbf{K}^{(t)}\in\mathbb{R}^{N\times d}$. Here, $N$ denotes the number of active agents, and $d$ is the embedding dimension of the semantic encoder. We posit that a communication link should exist from agent $j$ to agent $i$ if the semantic capacity offered by $j$ aligns with the need of $i$. We quantify semantic alignment using cosine similarity. We $\ell_2$-normalize embeddings and define
\begin{equation}
\label{eq:attention_score}
\scriptsize
\hat{\mathbf{q}}_i^{(t)}=\frac{\mathbf{q}_i^{(t)}}{\|\mathbf{q}_i^{(t)}\|_2},\quad
\hat{\mathbf{k}}_j^{(t)}=\frac{\mathbf{k}_j^{(t)}}{\|\mathbf{k}_j^{(t)}\|_2},\quad
r_{i,j}^{(t)}=(\hat{\mathbf{q}}_i^{(t)})^\top \hat{\mathbf{k}}_j^{(t)} \in [-1,1].
\end{equation}
This score measures how well agent $a_j$'s offered capability (key) matches agent $a_i$'s current need (query), and it is directly comparable across rounds under a fixed encoder.

\subsubsection{Sparse Graph Construction}
To obtain a sparse topology, we apply hard thresholding to the relevance matrix. The binary adjacency matrix $A^{(t)}\in\{0,1\}^{N\times N}$ is defined as
\begin{equation}
A_{j\rightarrow i}^{(t)}=\mathbb{I}\!\left(r_{i,j}^{(t)}>\tau_{\text{edge}}\right)\cdot(1-\delta_{ij}),
\end{equation}
where $\mathbb{I}(\cdot)$ is the indicator function, $\tau_{\text{edge}}$ controls graph sparsity, and $\delta_{ij}$ prevents self-loops.

We then set $\mathcal{E}^{(t)}=\{(a_j\rightarrow a_i)\mid A_{j\rightarrow i}^{(t)}=1\}$ and define the incoming neighbor set:
\begin{equation}
\mathcal{N}^{(t)}_{in}(i)=\{\,j\mid A_{j\rightarrow i}^{(t)}=1\,\}.
\end{equation}
Thus, a directed edge $a_j\rightarrow a_i$ indicates that $a_j$ is selected as a \emph{provider} for $a_i$ at round $t$, and $m_{priv,j}^{(t)}$ becomes eligible to be routed into $a_i$'s next-round context (Sec.~\ref{sec:state_update}).

\subsubsection{Topology Adaptation and Routing Semantics}
The induced topology $G^{(t)}=(\mathcal{A},\mathcal{E}^{(t)})$ is a sparse directed graph that encodes the instantaneous information dependencies of the collaboration at round $t$. Directionality is explicit: an edge $a_j\rightarrow a_i$ is activated only when $a_j$'s key embedding $\mathbf{k}_j^{(t)}$ semantically matches $a_i$'s query embedding $\mathbf{q}_i^{(t)}$, indicating that $a_j$ is selected as a provider for $a_i$ in the current round. This direction also determines routing semantics: when $a_j\rightarrow a_i$ is active, $m_{priv,j}^{(t)}$ becomes eligible to be routed into $a_i$'s next-round context (Sec.~\ref{sec:state_update}).

The topology is adaptive across rounds because the descriptors are recomputed from the evolving agent state. As the manager updates $C_{task}^{(t)}$ and agents update their memories $\{\mathcal{H}_i^{(t)}\}$, the textual descriptors $\{s^{(t)}_{q,i}, s^{(t)}_{k,i}\}$ change, which in turn changes the embeddings $\{\mathbf{q}_i^{(t)}, \mathbf{k}_i^{(t)}\}$ and the relevance scores $\{r_{i,j}^{(t)}\}$. Consequently, the adjacency matrix $A^{(t)}$ and edge set $\mathcal{E}^{(t)}$ can reconfigure from round to round, creating new links when a previously missing capability becomes relevant and removing links when an information need has been satisfied.

For example, an agent with a fixed \textit{Developer} role may shift its query from \textit{“need API specifications”} to \textit{“need test cases”} after implementing a module, while its key shifts from \textit{“can provide design draft”} to \textit{“can provide implementation code”}, leading to corresponding changes in its incoming and outgoing neighbors.

Sparsity acts as a practical communication budget control. The hard threshold $\tau_{\text{edge}}$ controls how many edges are activated, limiting irrelevant message traffic and reducing context overload from non-essential messages.

\subsection{Topology-Aware Message Ordering}
\label{sec:execution_scheduling}

DyTopo uses a synchronization barrier, so $G^{(t)}$ encodes routing eligibility rather than within-round execution causality. We therefore define deterministic message ordering only for prompt construction and trace reproducibility. For each recipient agent $a_i$, we order its incoming routed messages by decreasing semantic relevance $r^{(t)}_{i,j}$, breaking ties deterministically. This yields a recipient-specific order $\sigma_i^{(t)}$ used inside $\Sigma(\cdot)$ when updating $\mathcal{H}_i^{(t+1)}$.

\begin{table*}[t]
    \centering
    \caption{Statistics and characteristics of the evaluation datasets. We categorize datasets by domain and difficulty level to ensure a comprehensive evaluation of the multi-agent system.}
    \label{tab:datasets}
    \setlength{\tabcolsep}{7.5pt} 
    {\renewcommand{\arraystretch}{0.8}
    \begin{tabular}{l l l l} 
        \toprule
        \textbf{Dataset} & \textbf{Domain} & \textbf{Difficulty} & \textbf{Dataset Description} \\
        \midrule
        HumanEval & Code & Fundamental & Evaluation of functional correctness in Python via docstrings. \\
        APPS-Competition & Code & Advanced &  Competitive programming problems requiring algorithmic design. \\
        \midrule
        MATH-500 & Math & Hard & A balanced subset (7 domains) of MATH, focusing on derivation. \\
        Omni-MATH & Math & Expert & Focused on deep reasoning with Olympiad-level complexity. \\
        \bottomrule
    \end{tabular}
    }
\end{table*}

\subsubsection{Dependency Graph Definition}
At each round $t$, DyTopo induces an interaction graph
$G^{(t)}=(\mathcal{A},\mathcal{E}^{(t)})$ over the agent set $\mathcal{A}$. The edge set $\mathcal{E}^{(t)}$ is derived from the binary adjacency matrix $A^{(t)}$ constructed in Sec.~\ref{sec:dynamic_topology}, i.e., $\mathcal{E}^{(t)}=\{(a_j\rightarrow a_i)\mid A^{(t)}_{j\rightarrow i}=1\}$. A directed edge $a_j\rightarrow a_i$ represents an inference-time dependency: agent $a_i$ is selected to receive agent $a_j$'s private message in round $t$ (equivalently, $j\in\mathcal{N}^{(t)}_{in}(i)$), so $m^{(t)}_{priv,j}$ is eligible to be routed into $a_i$'s next-round context.

Given $G^{(t)}$, we define an aggregation order $\sigma^{(t)}=(\sigma^{(t)}_1,\dots,\sigma^{(t)}_N)$, a permutation of $\{1,\dots,N\}$, which is used to deterministically order routed messages in $\Sigma_{\sigma^{(t)}}(\cdot)$ during memory updates and to ensure reproducibility of the coordination trace. When multiple orders are possible, we break ties deterministically.

\subsubsection{Adaptive Topological Sequencing}
Because $G^{(t)}$ is induced dynamically, it may be acyclic or contain directed cycles. DyTopo constructs an aggregation order $\sigma^{(t)}=(\sigma^{(t)}_1,\dots,\sigma^{(t)}_N)$, a permutation of $\{1,\dots,N\}$, which is used to linearize dependencies when possible and provide a deterministic ordering for message integration via $\Sigma_{\sigma^{(t)}}(\cdot)$. For convenience, define the position of an agent index $i$ in the sequence as
\begin{equation}
\mathrm{pos}_{\sigma^{(t)}}(i)=\min\{\ell\in\{1,\dots,N\}\mid \sigma^{(t)}_\ell=i\}.
\end{equation}

\textbf{Case I: Directed Acyclic Graph (DAG).}
If $G^{(t)}$ contains no directed cycles, we compute $\sigma^{(t)}$ using a standard topological sort on $G^{(t)}$.
The resulting order satisfies
\begin{equation}
\forall\,(a_j\rightarrow a_i)\in\mathcal{E}^{(t)} \;\Rightarrow\;
\mathrm{pos}_{\sigma^{(t)}}(j) < \mathrm{pos}_{\sigma^{(t)}}(i),
\end{equation}
so that any selected provider $a_j$ precedes its consumer $a_i$ in the induced linearization. When multiple valid topological orders exist, we break ties to ensure reproducibility.

\textbf{Case II: Cyclic Graph.}
If $G^{(t)}$ contains cycles, no topological order exists. In this case, we construct $\sigma^{(t)}$ using a greedy
cycle-breaking heuristic based on the current dependency structure. Let $\mathcal{U}\subseteq\{1,\dots,N\}$ denote the set of unplaced agent indices, and define the restricted in-degree
\begin{equation}
d^{(t)}_{in}(i;\mathcal{U})=\left|\left\{j\in\mathcal{U}\mid (a_j\rightarrow a_i)\in\mathcal{E}^{(t)}\right\}\right|.
\end{equation}
We iteratively select the next index
\begin{equation}
i^*=\arg\min_{i\in\mathcal{U}} d^{(t)}_{in}(i;\mathcal{U}),
\end{equation}
append $i^*$ to $\sigma^{(t)}$, and remove it from $\mathcal{U}$ until all indices are placed. Intuitively, nodes with smaller restricted in-degree have fewer unmet dependencies within the remaining subgraph, so placing them earlier yields an order that minimizes reliance on information that is cyclically unavailable. This procedure produces a well-defined permutation even when $G^{(t)}$ is cyclic, and it provides a consistent ordering for
$\Sigma_{\sigma^{(t)}}(\cdot)$ in the memory update step.



\subsection{Meta-Control and Workflow Orchestration}
\label{sec:meta_control}

To prevent divergent discussions and ensure goal-oriented convergence, DyTopo includes a hierarchical control layer implemented by a Manager (meta-agent). Unlike worker agents $a_i\in\mathcal{A}$, which operate on local memories $\mathcal{H}_i^{(t)}$, the Manager maintains a global view of the current round and updates the shared round context $C_{task}^{(t)}$, which serves as the manager-specified round goal in Eq.~\eqref{eq:state_definition}.

\subsubsection{Global State Aggregation}
At the end of round $t$, after all worker agents have produced outputs under the synchronization barrier and private messages have been routed under $G^{(t)}$, the Manager aggregates the current-round public information into a global state representation
$S_{global}^{(t)}$. We define
\begin{equation}
S_{global}^{(t)} = 
\left[C_{task}^{(t)};\, \Sigma_{\sigma^{(t)}}\!\left(\{m_{pub,i}^{(t)} \mid a_i\in\mathcal{A}\}\right)\right],
\end{equation}
where $\Sigma_{\sigma^{(t)}}(\cdot)$ denotes structured, order-preserving concatenation under $\sigma^{(t)}$. This global view allows the Manager to track progress, detect inconsistencies, and identify missing information needed for convergence.

\begin{table*}[ht]
\centering
\caption{Multi-agent performance on coding and math benchmarks (\%, higher is better) across different LLM backbones.}
\label{tab:multi_agent_results}
\setlength{\tabcolsep}{4pt}
{\renewcommand{\arraystretch}{0.84}
\begin{tabular}{@{}lccccc@{}}
\toprule
\textbf{Backbone} & \textbf{LLM Output} & \textbf{Single-turn Agent} & \textbf{Random Topology} & \textbf{AgentScope} & \textbf{DyTopo}\\
\midrule

\multicolumn{6}{@{}l}{\textbf{HumanEval}}\\
\quad MiMo-V2-Flash & 86.59 & 88.41 (+2.1\%) & 88.17 (+1.8\%) & 90.24 (+4.2\%) & \textbf{92.07 (+6.3\%)} \\
\quad GPT-oss-120B      & 95.73 & 93.29 (-2.5\%) & 91.46 (-4.5\%) & 92.68 (-3.2\%) & \textbf{98.16 (+2.5\%)} \\
\quad Llama3-8B-Instruct & 23.78 & 47.56 (+100.0\%) & 42.07 (+77.0\%) & 42.94 (+80.6\%) & \textbf{50.61 (+112.9\%)} \\
\quad Qwen3-8B & 18.29 & 46.95 (+156.7\%) & 80.49 (+340.1\%) & 80.00 (+337.4\%) & \textbf{89.63 (+390.0\%)} \\
\midrule

\multicolumn{6}{@{}l}{\textbf{APPS-Competition}}\\
\quad MiMo-V2-Flash & 41.18 & 42.84 (+4.0\%) & 42.01 (+2.0\%) & 48.42 (+17.6\%) & \textbf{49.81 (+21.0\%)} \\
\quad GPT-oss-120B      & 31.88 & 60.55 (+90.0\%) & 52.21 (+63.8\%) & 58.71 (+84.2\%) & \textbf{69.66 (+118.5\%)} \\
\quad Llama3-8B-Instruct & 10.13 & 13.28 (+31.1\%) & 12.44 (+22.8\%) & 14.29 (+41.1\%) & \textbf{18.21 (+79.8\%)} \\
\quad Qwen3-8B & 9.62 & 13.47 (+40.0\%) & 24.24 (+152.0\%) & 23.12 (+140.3\%) & \textbf{25.14 (+161.3\%)} \\
\midrule

\multicolumn{6}{@{}l}{\textbf{Math-500}}\\
\quad MiMo-V2-Flash & 57.14 & 65.71 (+15.0\%) & 78.57 (+37.5\%) & 75.71 (+32.5\%) & \textbf{87.14 (+52.5\%)} \\
\quad GPT-oss-120B      & 60.00 & 81.43 (+35.7\%) & 87.14 (+45.2\%) & 82.86 (+38.1\%) & \textbf{88.57 (+47.6\%)} \\
\quad Llama3-8B-Instruct & 12.86 & 25.71 (+100.0\%) & 20.00 (+55.6\%) & 30.00 (+133.3\%) & \textbf{47.14 (+266.7\%)} \\
\quad Qwen3-8B & 48.57 & 57.14 (+17.6\%) & 68.57 (+41.2\%) & 72.86 (+50.0\%) & \textbf{75.71 (+55.9\%)} \\
\midrule

\multicolumn{6}{@{}l}{\textbf{Omni-Math}}\\
\addlinespace[2pt]
\quad MiMo-V2-Flash & 32.86 & 37.14 (+13.0\%) & 41.43 (+26.1\%) & 44.28 (+34.8\%) & \textbf{52.86 (+60.9\%)} \\
\quad GPT-oss-120B      & 14.29 & 25.71 (+80.0\%) & 34.29 (+140.0\%) & 37.14 (+160.0\%) & \textbf{41.43 (+190.0\%)} \\
\quad Llama3-8B-Instruct & 14.29 & 11.43 (-20.0\%) & 15.71 (+10.0\%) & 22.86 (+60.0\%) & \textbf{30.00 (+110.0\%)} \\
\quad Qwen3-8B & 21.43 & 32.86 (+53.3\%) & 25.71 (+20.0\%) & 35.71 (+66.6\%) & \textbf{51.43 (+140.0\%)} \\

\bottomrule
\end{tabular}
}

\medskip
\footnotesize
\textit{Note}: Relative improvement is computed per backbone. Bold indicates the best multi-agent result within each backbone row.
\end{table*}

\subsubsection{Manager Policy and Halting}
We model the Manager as a high-level policy $\Pi_{meta}$ that maps the global state to a halting decision and a next-round context update:
\begin{equation}
\langle y^{(t)},\, C_{task}^{(t+1)} \rangle \sim \Pi_{meta}(\cdot \mid S_{global}^{(t)}),
\end{equation}
where $y^{(t)}\in\{0,1\}$ indicates whether the system terminates after round $t$. Concretely, the halting decision is:
\begin{equation}
y^{(t)}=
\begin{cases}
1 & \text{if } \Phi(S_{global}^{(t)}) \ge \gamma_{success},\\
0 & \text{otherwise},
\end{cases}
\end{equation}
where $\Phi(\cdot)$ is an internal evaluation function and $\gamma_{success}$ is an acceptance threshold. When $y^{(t)}=0$, the updated context $C_{task}^{(t+1)}$ provides refined round-level guidance, focusing subsequent communication on unresolved subgoals.

\subsubsection{Bi-Level Feedback Loop}
If $y^{(t)}=0$, the Manager broadcasts $C_{task}^{(t+1)}$ to all agents and the system proceeds to round $t{+}1$ by updating the agent state in Eq.~\eqref{eq:state_definition}. This closes a feedback loop operating at two levels: at the micro-level, agents induce and use $G^{(t)}$ via query--key semantic matching (Sec.~\ref{sec:dynamic_topology}); at the macro-level, the Manager updates $C_{task}^{(t)}$ and decides when to halt.

\section{Experiments}

\subsection{Datasets}

To evaluate the reasoning and generation capabilities of our framework, we curate a benchmark suite spanning two domains: code generation and mathematical reasoning. The suite is intentionally difficulty-graded, ranging from function-level correctness to competition and Olympiad-level problem solving, to test whether the system remains reliable as task complexity and reasoning depth increase. We summarize these datasets in Table~\ref{tab:datasets}.

\textbf{Code generation benchmarks.}
We consider two programming benchmarks with complementary difficulties. HumanEval~\citep{chen2021evaluatinglargelanguagemodels} provides a baseline assessment of fundamental coding ability through handwritten Python problems that primarily test docstring understanding and function-level correctness. To evaluate performance under substantially higher complexity, we additionally use a subset of APPS~\citep{hendrycksapps2021}, selecting 100 problems from the Competition split. These tasks resemble collegiate programming contests and typically require robust handling of edge cases and more sophisticated algorithmic design.

\textbf{Mathematical reasoning benchmarks.}
For mathematical reasoning, we target multi-step deduction across diverse subfields. We use MATH-500~\citep{lightman2023lets} as a representative benchmark to probe consistent multi-step reasoning. We also include Omni-MATH~\citep{gao2024omnimathuniversalolympiadlevel}, a dataset intended to evaluate long-horizon reasoning behavior under complex, multi-step solution trajectories. We sample 70 Omni-MATH problems using the same seven-domain, 10-per-domain stratification. Compared with MATH-500, Omni-MATH more frequently requires extended Olympiad-level solution trajectories, making it a stringent test of maintaining coherent reasoning over longer contexts.

\subsection{Settings}


DyTopo is model-agnostic and can instantiate each agent with different LLM backbones. We evaluate the framework using both proprietary and open-weights models, including mimo-v2-flash~\citep{xiao2026mimo}, GPT-oss-120B~\citep{agarwal2025gpt}, Llama-3-8B-Instruct~\citep{grattafiori2024llama} and Qwen3-8B~\citep{yang2025qwen3}, served via vLLM~\citep{kwon2023efficientmemorymanagementlarge}. For the semantic encoding, we use the sentence embedding model all-MiniLM-L6-v2 and compute cosine similarity between agent profiles.

Our experiments cap interaction at $T_{\max}$ rounds, but enable Manager-controlled early stopping by default (Sec.~\ref{sec:meta_control}): the Manager halts as soon as task-specific completion criteria are met. Unless otherwise stated, DyTopo therefore runs a variable number of rounds per instance with a hard cap of $T_{\max}$. In the communication-round ablation (Sec.~\ref{sec:rounds_ablation}), we disable halting and force exactly $T$ rounds to isolate the effect of interaction depth. We provide the complete prompt and more detailed experimental parameters in Appendix~\ref{app:implementation}.

\section{Results}

We report four sets of results: (i) main benchmark performance across methods and backbones, (ii) the effect of communication rounds, (iii) qualitative analysis of dynamic topology evolution, and (iv) ablations on the semantic similarity threshold used for Q-K routing. Additionally, we have included Token Usage and Latency Analysis in Appendix~\ref{app:efficiency}.

\subsection{Main results across benchmarks}
\label{sec:main_results}

\Cref{tab:multi_agent_results} reports accuracy on code generation (HumanEval, APPS-Competition) and mathematical reasoning (Math-500, Omni-Math) across four LLM backbones. DyTopo is the best method in all 16 backbone$\times$dataset settings, improving over the strongest non-DyTopo baseline by 0.90-17.14 points (mean +6.09). These consistent wins indicate that the benefit is not tied to a particular model family or task, but to round-adaptive, content-driven routing.

Gains are especially pronounced on the harder math benchmarks, where selective verification and error localization matter most. On Math-500, DyTopo improves by up to +17.14 (Llama3-8B: 47.14 vs.\ 30.00), and on Omni-Math by up to +15.72 (Qwen3-8B: 51.43 vs.\ 35.71). On code generation, DyTopo also yields reliable improvements, including +9.11 on APPS-Competition with GPT-oss-120B (69.66 vs.\ 60.55) and +9.14 on HumanEval with Qwen3-8B (89.63 vs.\ 80.49).

Multi-round interaction alone is not sufficient: Random Topology can help in some cases but is inconsistent, whereas DyTopo improves uniformly. The following analyses connect these gains to (i) non-monotonic returns with more rounds, (ii) a stage-wise shift from exploratory to verification/assembly topologies, and (iii) a sparsity ``sweet spot'' controlled by the similarity threshold.

\subsection{Effect of communication rounds}
\label{sec:rounds_ablation}

\begin{figure}[t]
\centering
\includegraphics[width=\linewidth]{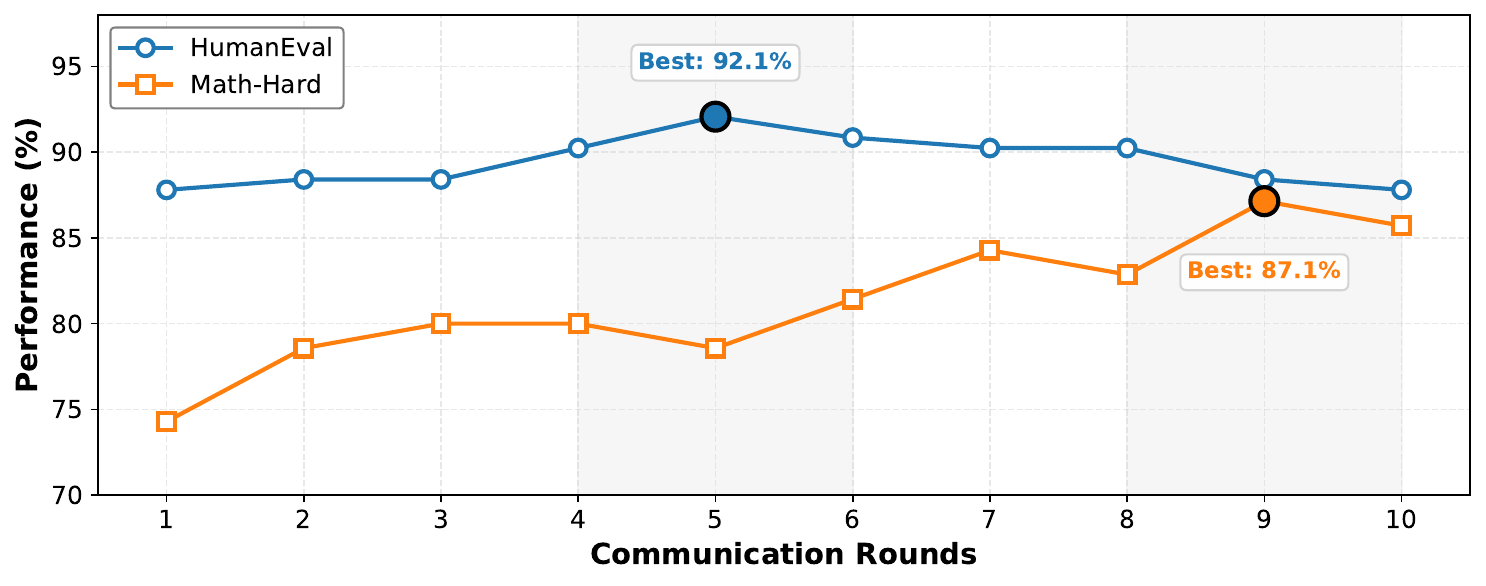}
\caption{Performance change on communication rounds for HumanEval and Math-500. HumanEval achieves optimal performance (92.07\%) at the 5th round, while Math-500 peaks (87.14\%) at the 9th round, suggesting task-specific performances.}
\label{fig:rounds_ablation}
\end{figure}

We then study how performance changes as we vary the number of communication rounds while keeping the agent pool and routing mechanism fixed. \Cref{fig:rounds_ablation} shows a non-monotonic trend on both a coding benchmark (HumanEval) and a math benchmark (Math-500), indicating that more rounds do not always help.

HumanEval peaks at 5 rounds (92.07\%), after which additional rounds slightly degrade performance, consistent with the hypothesis that once a correct implementation is reached, extra communication can introduce unnecessary edits or distractors. In contrast, Math-500 continues improving for longer and peaks at 9 rounds (87.14\%), suggesting that difficult mathematical reasoning benefits from extended iterative refinement, verification, and error correction. Overall, these results highlight that the optimal communication budget is task-dependent, motivating a manager-driven process that can adapt the trajectory (and potentially stopping behavior) based on round-by-round progress rather than relying on a single fixed round count for all tasks.

\begin{figure*}[t]
\centering
\includegraphics[width=0.91\linewidth]{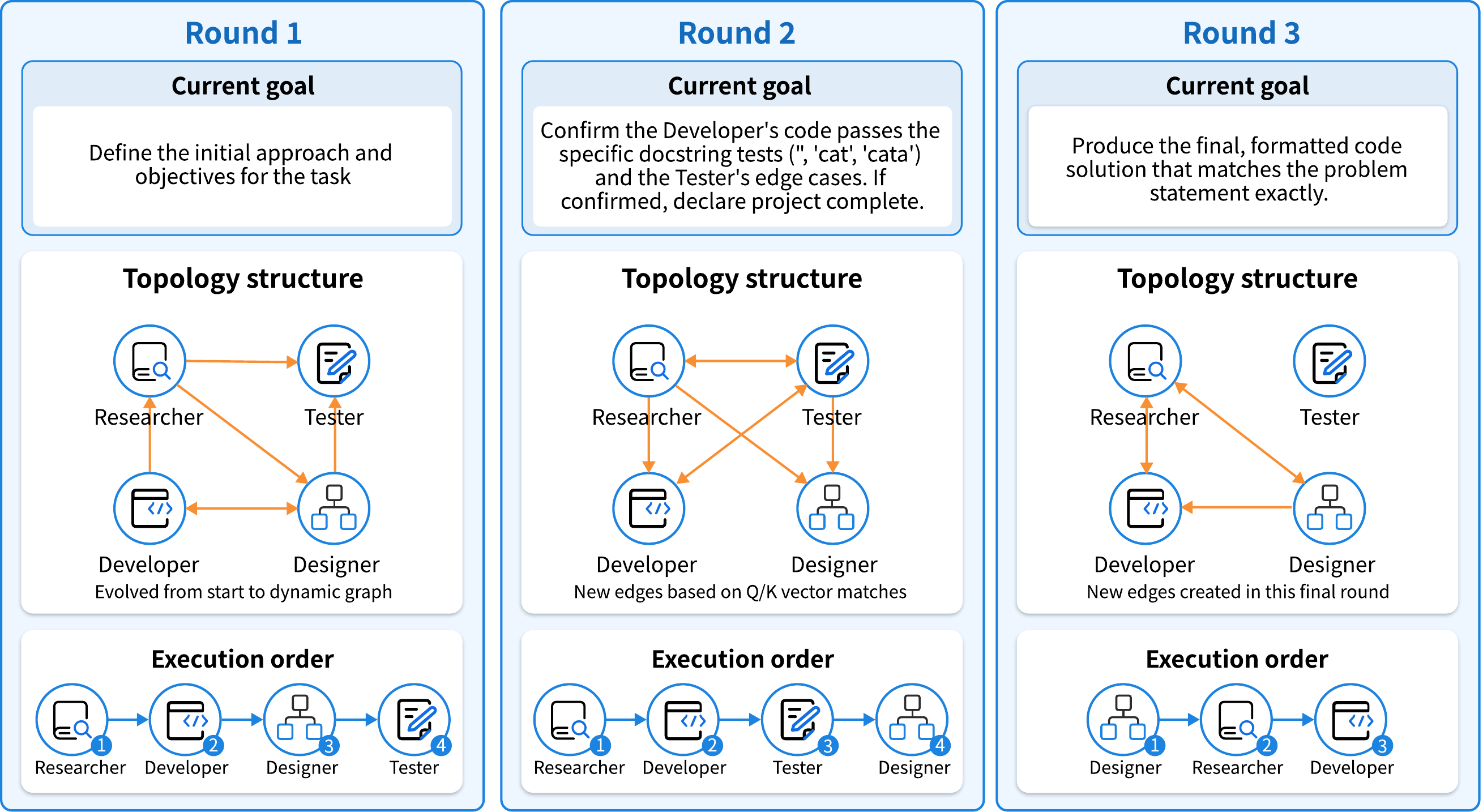}
\caption{\textbf{DyTopo rewires communication over rounds.} For one code-generation instance, we show three representative rounds (goal, induced directed graph $G^{(t)}$, and execution order). The topology transitions from broad, exploratory routing (Round 1) to verification-focused connections (Round 2), and finally to a sparse, dependency-minimal graph for producing the formatted final answer (Round 3).}
\label{fig:topology_evolution}
\end{figure*}

\subsection{Topology evolution and interpretability}
\label{sec:topology_evolution_results}

DyTopo exposes an explicit coordination trace through the round-wise induced graphs $\{G^{(t)}\}$, making multi-agent interaction qualitatively analyzable rather than implicit. Figure~\ref{fig:topology_evolution} shows that the topology reorganizes in a goal-aligned manner on a representative HumanEval instance: the early round exhibits higher edge density and broader reach, consistent with exploratory problem framing and decomposition, where multiple roles exchange partial constraints and candidate strategies.

As the round goal shifts toward verification, the communication pattern becomes more selective and feedback-driven. The induced edges concentrate around agents whose outputs operationalize checking and those responsible for incorporating corrections. The accompanying aggregation order reinforces this interpretation by prioritizing diagnostic signals before consolidation, suggesting that DyTopo is not merely changing \emph{who} talks, but also \emph{when} information is routed to affect downstream decisions.

In the final round, the topology prunes to a dependency-minimal subgraph oriented toward assembling the formatted solution, reflecting reduced uncertainty and fewer outstanding queries. This stage-wise sparsification highlights which agent pairs are structurally critical at each stage and offers a concrete debugging handle.

Additionally, we provide a more comprehensive case in Appendix~\ref{app:qualitative} to aid in understanding topological evolution and the Q-K matching mechanism.

\subsection{Ablation on Q-K similarity threshold}
\label{sec:threshold_ablation}

Finally, we ablate the similarity threshold $\tau$ used in the hard-threshold adjacency construction (Sec.~\ref{sec:dynamic_topology}). As shown in Table~\ref{tab:threshold_ablation_compact}, performance exhibits a clear optimum as $\tau$ controls the sparsity of $G^{(t)}$.
\begin{table}[tbp]
  \centering
  \caption{Ablation study on similarity threshold across different datasets. The performance metrics (in \%) demonstrate the sensitivity of the method to this hyperparameter. Best results for each dataset are highlighted in bold.}
  \label{tab:threshold_ablation_compact}
  \setlength{\tabcolsep}{2pt} 
  \renewcommand{\arraystretch}{1.2}
  \resizebox{\columnwidth}{!}{%
      \begin{tabular}{@{}l *{9}{c}@{}}
        \toprule
        \multirow{2}{*}{\textbf{Dataset}} & 
        \multicolumn{9}{c}{\textbf{Similarity Threshold}} \\
        \cmidrule(l){2-10}
        & 0.1 & 0.2 & 0.3 & 0.4 & 0.5 & 0.6 & 0.7 & 0.8 & 0.9 \\
        \midrule
        apps-comp. & 43.61 & 44.84 & \textbf{49.81} & 47.35 & 44.13 & 40.86 & 40.47 & 37.59 & 34.51 \\
        omni\_math & 42.86 & 47.14 & 50.00 & \textbf{52.86} & 48.57 & 42.86 & 41.43 & 40.00 & 38.58 \\
        \bottomrule
        \vspace{0pt}
      \end{tabular}%
  }
  \footnotesize\textit{Note: Performance metrics (in \%) show optimal thresholds vary by dataset (0.3 for apps-competition, 0.4 for omni\_math).}
\end{table}

For APPS-Competition, the best result occurs at $\tau = 0.3$ (49.81\%), while for Omni-Math the optimum shifts higher to $\tau = 0.4$ (52.86\%). Two failure modes are apparent at the extremes: when $\tau$ is low, the topology becomes overly dense, increasing irrelevant message traffic and degrading effective context utilization; when $\tau$ is too high, the topology becomes sparse, preventing useful information flow and reducing collaboration benefits. The fact that the optimal $\tau$ differs across datasets further supports the conclusion that communication structure is task-sensitive, and that controlling sparsity is a key practical knob for stable performance.

\section{Conclusion}
We presented \textbf{DyTopo}, a multi-agent framework that dynamically rewires a sparse directed communication graph at each round via semantic matching between agents’ natural-language \emph{query} (need) and \emph{key} (offer) descriptors. This inference-time routing aligns information flow with stage-dependent goals, yielding both stronger coordination and an interpretable coordination trace through the evolving graphs. Across code generation and mathematical reasoning benchmarks and multiple LLM backbones, DyTopo consistently outperforms fixed and random-topology baselines, while ablations show communication budget and sparsity are task-sensitive.

\clearpage

\section*{Impact Statement}
This paper introduces DyTopo, a method that improves multi-agent LLM collaboration by dynamically routing messages through sparse, goal-conditioned communication graphs induced via semantic matching of lightweight need/offer descriptors. The intended impact is to advance reliable and interpretable multi-round reasoning, with potential benefits for applications such as code generation and mathematical problem solving. Potential risks include misuse of improved agentic capabilities and privacy concerns if agent messages or coordination traces are logged in sensitive settings. DyTopo can also fail when descriptors are misleading, causing misrouting and error propagation. We recommend pairing DyTopo with standard safety filters, secure logging practices, and application-specific guardrails when deployed.

\nocite{langley00}

\bibliography{example_paper}
\bibliographystyle{icml2026}

\newpage
\appendix
\onecolumn

\section{Complexity Analysis}
\label{app:algorithm}

\subsection{Computational Complexity}
In this section, we analyze the computational complexity of DyTopo compared to static fully-connected topologies. Let $N$ be the number of agents.

\textbf{Communication Cost.} In a standard fully-connected multi-agent system, the message passing complexity is $O(N^2)$ per round, as every agent attends to every other agent's output. DyTopo introduces a dynamic sparsity mechanism. The graph construction involves computing $N^2$ pairwise similarities, which has a complexity of $O(N^2)$. However, the actual message routing is sparse. Let $\rho$ be the sparsity factor determined by the threshold $\tau$, such that the number of active edges $|\mathcal{E}^{(t)}| \approx \rho N^2$ where $\rho \ll 1$. The subsequent context processing cost for the LLM is significantly reduced because agents only ingest messages from relevant neighbors defined by $\mathcal{N}_{in}^{(t)}$, avoiding context window overflow and reducing irrelevant distraction.

\textbf{Inference Overhead.} The overhead introduced by the Semantic Matching Engine (embedding generation and dot-product) is negligible compared to the LLM inference time. Using the \texttt{all-MiniLM-L6-v2} model (approx. 22M parameters), the embedding latency is in the order of milliseconds, whereas LLM generation (e.g., MiMo-V2-Flash or GPT-oss-120B) takes seconds. Thus, DyTopo improves reasoning performance without introducing a significant latency bottleneck.

\section{Implementation Details}
\label{app:implementation}

\subsection{System Prompts and Templates}
\label{app:prompts}

To ensure reproducibility, we provide the exact system prompts used for the Manager and Worker Agents across different domains. All prompts are designed to enforce strict JSON output formats to facilitate the extraction of Key ($s_k$) and Query ($s_q$) descriptors.

\subsubsection{Code Generation Agents}
For code generation tasks, the system utilizes a Manager and four specialized worker agents: Developer, Researcher, Tester, and Designer.

\textbf{Manager Agent (Code Generation).}
The Manager orchestrates the workflow and determines task completion based on the existence of code and passing tests.

\begin{tcolorbox}[title=Manager Agent Prompt (Code Gen), colback=gray!5, colframe=gray!40]
\small
\texttt{You are Manager, the Workflow Orchestrator. Decide if the task is COMPLETE.} \\
\texttt{STRICT RESPONSE FORMAT - MANDATORY} \\
\texttt{You MUST output ONLY a valid JSON object with these exact fields:} \\
\texttt{\{} \\
\texttt{  "public\_content": "String. Status summary and next directives.",} \\
\texttt{  "private\_content": \{\}, // No successors for manager} \\
\texttt{  "q\_desc": "String. REQUIRED. What you need next (query descriptor).",} \\
\texttt{  "k\_desc": "String. REQUIRED. What you provide (key descriptor).",} \\
\texttt{  "is\_complete": Boolean,} \\
\texttt{  "next\_goal": "String"} \\
\texttt{\}} \\
\texttt{CRITICAL RULES:} \\
\texttt{1. Primary focus: Workflow Orchestrator. Decide if the task is COMPLETE.} \\
\texttt{2. Strict constraint: Only set is\_complete=True if Code exists AND Tests pass.} \\
\texttt{3. If you see Python code, analyze it from your role's perspective.}
\end{tcolorbox}

\textbf{Worker Agents (Code Generation).}
All worker agents share a common prompt template structure but utilize distinct role descriptions.

\begin{tcolorbox}[title=Generic Worker Agent Prompt Template, colback=gray!5, colframe=gray!40]
\small
\texttt{You are [Agent Name], the [Agent Role].} \\
\texttt{Role Description: [Insert Role Specific Description]} \\
\texttt{STRICT RESPONSE FORMAT - MANDATORY} \\
\texttt{You MUST output ONLY a valid JSON object with these exact fields:} \\
\texttt{\{} \\
\texttt{  "public\_content": "String. Your contribution to the task.",} \\
\texttt{  "private\_content": \{"TargetRole": "Instruction"\},} \\
\texttt{  "q\_vector": "String. REQUIRED. What you need next (Query).",} \\
\texttt{  "k\_vector": "String. REQUIRED. What you provide (Key)."} \\
\texttt{\}}
\end{tcolorbox}

\textbf{Role Descriptions.}
Table \ref{tab:code_roles} details the specific responsibilities injected into the template for each worker role.

\subsubsection{Mathematical Reasoning Agents}
For mathematical reasoning tasks, we employ a specialized set of agents: ProblemParser, Solver, Verifier, and Manager.

\textbf{Role Definitions.}
Table \ref{tab:math_roles} outlines the responsibilities and constraints for each agent in the mathematical domain.

\textbf{System Prompts.}
Below are the exact system prompts used for the math agents.

\begin{tcolorbox}[title=ProblemParser Agent Prompt, colback=gray!5, colframe=gray!40]
\small
\texttt{You are [Agent Name], the Math Problem Analyst. Analyze the problem statement, identify known conditions, define the solving target, and break down the solving steps.} \\
\texttt{STRICT RESPONSE FORMAT - MANDATORY} \\
\texttt{You MUST output ONLY a valid JSON object with these exact fields:} \\
\texttt{\{} \\
\texttt{  "public\_content": "String. Analysis, conditions, target, and plan.",} \\
\texttt{  "private\_content": \{"Role": "Instruction"\},} \\
\texttt{  "q\_vector": "String. REQUIRED. What you need next.",} \\
\texttt{  "k\_vector": "String. REQUIRED. What you provide."} \\
\texttt{\}} \\
\texttt{CRITICAL RULES:} \\
\texttt{1. Primary focus: Analyze problem, identify conditions, define target, break down steps.} \\
\texttt{2. Constraint: MAX 1000 words. Output MUST include clear analysis and plan.} \\
\texttt{3. Analyze mathematical expressions from your role's perspective.}
\end{tcolorbox}

\begin{tcolorbox}[title=Solver Agent Prompt, colback=gray!5, colframe=gray!40]
\small
\texttt{You are [Agent Name], the Mathematical Solver. Execute specific mathematical calculations, symbolic reasoning, or theorem calls. Can use SymPy for symbolic computations.} \\
\texttt{STRICT RESPONSE FORMAT - MANDATORY} \\
\texttt{You MUST output ONLY a valid JSON object with these exact fields:} \\
\texttt{\{} \\
\texttt{  "public\_content": "String. Detailed solutions with derivations.",} \\
\texttt{  "private\_content": \{"Role": "Instruction"\},} \\
\texttt{  "answer": "String. The final answer.",} \\
\texttt{  "q\_vector": "String. REQUIRED.",} \\
\texttt{  "k\_vector": "String. REQUIRED."} \\
\texttt{\}} \\
\texttt{CRITICAL RULES:} \\
\texttt{1. Primary focus: Calculations, symbolic reasoning, theorem calls.} \\
\texttt{2. Constraint: MAX 2000 words. Detailed step-by-step solutions required.}
\end{tcolorbox}

\begin{tcolorbox}[title=Verifier Agent Prompt, colback=gray!5, colframe=gray!40]
\small
\texttt{You are [Agent Name], the Logic Verifier. Check the rationality of each derivation step, identify logical loopholes or calculation errors.} \\
\texttt{STRICT RESPONSE FORMAT - MANDATORY} \\
\texttt{You MUST output ONLY a valid JSON object with these exact fields:} \\
\texttt{\{} \\
\texttt{  "public\_content": "String. Verification results and conclusion.",} \\
\texttt{  "private\_content": \{"Role": "Instruction"\},} \\
\texttt{  "q\_vector": "String. REQUIRED.",} \\
\texttt{  "k\_vector": "String. REQUIRED."} \\
\texttt{\}} \\
\texttt{CRITICAL RULES:} \\
\texttt{1. Primary focus: Check rationality, find loopholes/errors.} \\
\texttt{2. Constraint: MAX 1000 words. Output verification for each step.}
\end{tcolorbox}

\begin{tcolorbox}[title=Manager Agent Prompt (Math), colback=gray!5, colframe=gray!40]
\small
\texttt{You are [Agent Name], the Workflow Orchestrator. Decide if the task is COMPLETE.} \\
\texttt{STRICT RESPONSE FORMAT - MANDATORY} \\
\texttt{You MUST output ONLY a valid JSON object with these exact fields:} \\
\texttt{\{} \\
\texttt{  "public\_content": "String. Status summary.",} \\
\texttt{  "private\_content": \{"Role": "Instruction"\},} \\
\texttt{  "q\_vector": "String.", "k\_vector": "String.",} \\
\texttt{  "is\_complete": Boolean, "next\_goal": "String"} \\
\texttt{\}} \\
\texttt{CRITICAL RULES:} \\
\texttt{1. Strict constraint: Only set is\_complete=True if Solution exists AND Verification passes.}
\end{tcolorbox}

\textbf{Semantic Descriptors (Key/Query).}
To facilitate dynamic routing, each math agent generates domain-specific query ($s_q$) and key ($s_k$) textual descriptors, which are later embedded into vectors by the semantic encoder.

\begin{table}[h]
\centering
\caption{Agent Role Descriptions for Code Generation Tasks.}
\label{tab:code_roles}
\setlength{\tabcolsep}{16pt}
\renewcommand{\arraystretch}{1.2}
\begin{tabular}{@{}l l@{}}
\toprule
\textbf{Role} & \textbf{Description} \\ \midrule
\textbf{Developer} & Implement complete, runnable code. If using classes, provide independent functions as entry points. \\
\textbf{Researcher} & Identify standard algorithms and time complexity. Output conclusions without derivation. \\
\textbf{Tester} & Provide critical test cases and expected results. Describe testing logic rather than full execution logs. \\
\textbf{Designer} & Design API interfaces and class structures. Only show method signatures and type hints. \\ \bottomrule
\end{tabular}%
\end{table}

\begin{table}[h]
\centering
\caption{Agent Roles and Responsibilities for Mathematical Reasoning.}
\label{tab:math_roles}
\setlength{\tabcolsep}{4pt}
\renewcommand{\arraystretch}{1.2}
\begin{tabular}{@{}l l p{9cm}@{}@{}}
\toprule
\textbf{Role} & \textbf{Primary Responsibility} & \textbf{Core Constraints} \\ \midrule
\textbf{ProblemParser} & Decompose the problem statement. & Output must include analysis, known conditions, target, and a step-by-step plan. \\
\textbf{Solver} & Execute mathematical derivation. & Provide detailed steps, symbolic reasoning, and the final answer. \\
\textbf{Verifier} & Logic and calculation check. & Verify the rationality of each step; identify logical loopholes or calculation errors. \\
\textbf{Manager} & Workflow orchestration. & Halt only when a solution exists AND verification passes. \\ \bottomrule
\end{tabular}%
\end{table}

\begin{table}[h]
\centering
\caption{Representative query ($s_q$) and key ($s_k$) textual descriptors for mathematical agents (embedded later for routing).}
\label{tab:math_qk}
\setlength{\tabcolsep}{4pt}
\renewcommand{\arraystretch}{1.2}
\begin{tabular}{@{}l l p{7.5cm}@{}@{}}
\toprule
\textbf{Role} & \textbf{Typical Query Descriptor ($s_q$)} & \textbf{Typical Key Descriptor ($s_k$)} \\ \midrule
\textbf{ProblemParser} & I need the problem statement to analyze. & I provide problem analysis and solving plan. \\
\textbf{Solver} & I need a problem analysis and solving plan. & I provide detailed mathematical solutions with step-by-step derivations. \\
\textbf{Verifier} & I need a detailed solution to verify. & I provide verification of mathematical solutions. \\
\textbf{Manager} & I need status updates from all team members. & I coordinate team efforts and set priorities. \\ \bottomrule
\end{tabular}%
\end{table}

\subsection{Model Configurations}
\textbf{Embedding Model.} For the semantic matching mechanism, we utilize \texttt{sentence-transformers/all-MiniLM-L6-v2} (Hugging Face). This model maps the natural language $s_q$ and $s_k$ descriptors into a 384-dimensional vector space.

\textbf{Baselines.}
\begin{itemize}
    \item \textbf{Random Topology:} Implemented by enforcing the same sparsity level as DyTopo but randomizing the edge connections at each round. This controls for the effect of graph sparsity, isolating the contribution of \textit{semantic} routing.
    \item \textbf{Static Topology:} A fixed graph structure is predefined and reused across all communication rounds.
    \item \textbf{AgentScope:} A standard pipeline-based multi-agent framework where communication follows a fixed sequential order and a central hub pattern, without dynamic rewiring based on content.
\end{itemize}

\begin{table}[t]
    \centering
    \caption{Hyperparameters for the experiments.}
    \label{tab:hyperparams}
    \setlength{\tabcolsep}{13pt}
    \renewcommand{\arraystretch}{1.2}
    \begin{tabular}{l c c l}
    \toprule
    \textbf{Category} & \textbf{Symbol} & \textbf{Value} & \textbf{Description} \\
    \midrule
    \multicolumn{4}{l}{\textit{General Configuration}} \\
    Agents & $N$ & 4-6 & Size of the agent pool. \\
    Rounds & $T$ & 10 & Maximum interaction rounds. \\
    \midrule
    \multicolumn{4}{l}{\textit{Topology Evolution}} \\
    Similarity threshold & $\tau_{\text{edge}}$ & 0.1--0.9 & Minimum cosine similarity to activate an edge. \\
    \midrule
    \multicolumn{4}{l}{\textit{Generation Configuration}} \\
    Temperature & $T_{\text{gen}}$ & 0.3 & LLM decoding temperature for agent generation. \\
    Max tokens & $L_{\max}$ & 3000--5000 & Upper bound on generated tokens per response. \\
    Max in-degree & $K_{\text{in}}$ & 3 & Max number of providers routed into each agent per round. \\
    JSON enforcement & -- & True & Constrain outputs to structured JSON when required. \\
    \bottomrule
    \end{tabular}
\end{table}

\section{Pseudo-code for DyTopo}
\label{sec:Pseudo-code}

\begin{algorithm}[h]
\caption{DyTopo: Dynamic Topology Routing via Semantic Matching}
\label{alg:dynamic_reasoning}
\begin{algorithmic}[1]
\STATE {\bfseries Input:} Agent set $\mathcal{A}$, initial task context $C_{task}^{(0)}$, max rounds $T_{\max}$
\STATE {\bfseries Initialize:} $\mathcal{H}_i^{(0)} \leftarrow \emptyset$ for all $a_i\in\mathcal{A}$; $t\leftarrow 0$; $y\leftarrow 0$
\WHILE{$t < T_{\max}$ \AND $y=0$}
    \STATE \textbf{Phase 1: Single-Pass Agent Inference (Sec.~\ref{sec:methods})}
    \FORALL{$a_i \in \mathcal{A}$}
        \STATE $S_i^{(t)} \leftarrow [\rho_i;\, C_{task}^{(t)};\, \mathcal{H}_i^{(t)}]$
        \STATE $O_i^{(t)}=\langle m^{(t)}_{pub,i},\, m^{(t)}_{priv,i},\, s^{(t)}_{q,i},\, s^{(t)}_{k,i}\rangle \sim \pi_{\theta_i}(\cdot \mid S_i^{(t)})$
    \ENDFOR

    \STATE \textbf{Phase 2: Topology Induction (Sec.~\ref{sec:dynamic_topology})}
    \FORALL{$a_i \in \mathcal{A}$}
        \STATE $\mathbf{q}_i^{(t)} \leftarrow \mathrm{Emb}(s^{(t)}_{q,i}),\quad \mathbf{k}_i^{(t)} \leftarrow \mathrm{Emb}(s^{(t)}_{k,i})$
        \STATE $\hat{\mathbf{q}}_i^{(t)} \leftarrow \mathbf{q}_i^{(t)}/\|\mathbf{q}_i^{(t)}\|_2,\quad \hat{\mathbf{k}}_i^{(t)} \leftarrow \mathbf{k}_i^{(t)}/\|\mathbf{k}_i^{(t)}\|_2$
    \ENDFOR
    \STATE \textbf{Compute relevance matrix:} $R^{(t)} \in \mathbb{R}^{N\times N}$ where $R^{(t)}_{i,j} \leftarrow (\hat{\mathbf{q}}_i^{(t)})^\top \hat{\mathbf{k}}_j^{(t)}$
    \STATE \textbf{Threshold to adjacency:} $A^{(t)}_{j\rightarrow i} \leftarrow \mathbb{I}(R^{(t)}_{i,j}>\tau_{\text{edge}})\cdot(1-\delta_{ij})$ for all $i,j$
    \STATE $G^{(t)}\leftarrow(\mathcal{A},\mathcal{E}^{(t)})$ where $\mathcal{E}^{(t)}=\{(a_j\!\rightarrow\! a_i)\mid A_{j\rightarrow i}^{(t)}=1\}$

    \STATE \textbf{Phase 3: Deterministic Message Ordering (Sec.~\ref{sec:execution_scheduling})}
    \STATE Construct an order $\sigma^{(t)}$ (e.g., topological order if acyclic; otherwise a deterministic heuristic)

    \STATE \textbf{Phase 4: Routing \& Memory Update (Sec.~\ref{sec:state_update})}
    \FORALL{$a_i \in \mathcal{A}$}
        \STATE $\mathcal{N}^{(t)}_{in}(i)\leftarrow \{j \mid (a_j\rightarrow a_i)\in\mathcal{E}^{(t)}\}$
        \STATE $\mathcal{H}_i^{(t+1)} \leftarrow \mathcal{H}_i^{(t)} \oplus m^{(t)}_{pub,i} \oplus \Sigma_{\sigma^{(t)}}(\{m^{(t)}_{priv,j}\mid j\in\mathcal{N}^{(t)}_{in}(i)\})$
    \ENDFOR

    \STATE \textbf{Phase 5: Manager Control (Sec.~\ref{sec:meta_control})}
    \STATE $S_{global}^{(t)} \leftarrow [C_{task}^{(t)};\, \Sigma_{\sigma^{(t)}}(\{m^{(t)}_{pub,i}\mid a_i\in\mathcal{A}\})]$
    \STATE $\langle y,\, C_{task}^{(t+1)}\rangle \sim \Pi_{meta}(\cdot \mid S_{global}^{(t)})$
    \STATE $t \leftarrow t+1$
\ENDWHILE
\STATE {\bfseries Output:} final solution extracted from the last global state
\end{algorithmic}
\end{algorithm}

\section{Additional Experimental Results}
\subsection{Token Usage and Latency Analysis}
\label{app:efficiency}

We conduct a detailed breakdown of computational costs and inference latency using the \texttt{mimo-v2-flash} backbone on the HumanEval benchmark. To rigorously evaluate efficiency, we compare DyTopo against four baselines with distinct structural configurations:

\begin{itemize}
    \item \textbf{LLM Output:} A standard single-agent, single-pass generation (1 Agent, 1 Round).
    \item \textbf{Single-turn Agent:} An ensemble-like approach where 4 worker agents generate solutions in parallel without communication (4 Agents, 1 Round).
    \item \textbf{Random Topology:} A multi-agent system with random connectivity, forced to run for a fixed horizon (4 Agents, 5 Rounds).
    \item \textbf{AgentScope:} A standard ReAct-based multi-agent framework using a broadcast topology, also running for a fixed horizon (4 Agents, 5 Rounds).
\end{itemize}

\textbf{Analysis of Token Consumption.}
Table \ref{tab:token_latency} reveals the trade-offs between interaction depth and resource cost.
\begin{itemize}
    \item \textbf{The Cost of Fixed Horizons (AgentScope \& Random):} The \textit{AgentScope} baseline consumes the most resources (19,520 tokens). This bloat stems from two factors: (1) the fixed 5-round trajectory prevents early stopping even after a solution is found, and (2) the ReAct-style reasoning within each agent generates significant intermediate "thought" tokens. Similarly, \textit{Random Topology} incurs high costs (15,783 tokens) purely due to the fixed 5-round duration, yet achieves lower accuracy (88.17\%) than the single-turn ensemble (88.41\%), proving that prolonged interaction without semantic guidance introduces noise.
    \item \textbf{Efficiency via Fast Convergence (DyTopo):} \textbf{DyTopo} achieves the highest accuracy (\textbf{92.07\%}) while consuming only \textbf{48\%} of the tokens required by AgentScope (9,453 vs. 19,520). This efficiency is driven by our \textit{Manager-controlled halting mechanism}. As shown in the "Avg Rounds" column, DyTopo typically converges to a correct solution in just \textbf{2-3 rounds} (avg 2.6). By dynamically halting the conversation once the \textit{Verifier or Tester} confirms correctness, DyTopo avoids the redundant computations that plague fixed-horizon baselines.
\end{itemize}

\textbf{Latency Comparison.}
Latency follows a similar trend. \textit{Single-turn} methods are fastest due to parallel execution. Among multi-round systems, DyTopo (22.3s) is significantly faster than AgentScope (39.8s). The reduction in wall-clock time is a direct result of processing fewer communication rounds and using a sparse dependency graph, which reduces the context length for each inference call.

\begin{table}[h]
\centering
\caption{Token Usage and Latency Analysis on HumanEval. Config denotes (worker agents $\times$ Avg Rounds); unless stated otherwise, a separate Manager is invoked once per round for all multi-round methods. DyTopo achieves state-of-the-art accuracy with significantly lower costs than other multi-round baselines by converging rapidly (in $\sim$2.6 rounds) and utilizing sparse routing, whereas AgentScope and Random Topology suffer from fixed-horizon overheads.}
\label{tab:token_latency}
\setlength{\tabcolsep}{13pt}
\renewcommand{\arraystretch}{1.2}
\begin{tabular}{@{}l c c c c c@{}}
\toprule
\textbf{Method} & \textbf{Config} & \textbf{Accuracy} & \textbf{Total Tokens} & \textbf{Avg Latency (s)} & \textbf{Cost vs. DyTopo} \\ \midrule
LLM Output & $1 \times 1$ & 86.59 & 635 & 1.5 & $0.07\times$ \\
Single-turn Agent & $4 \times 1$ & 88.41 (+2.1\%) & 2,835 & 6.7 & $0.30\times$ \\ \midrule
Random Topology & $4 \times 5$ & 88.17 (+1.8\%) & 15,783 & 34.2 & $1.67\times$ \\
AgentScope & $4 \times 5$ & 90.24 (+4.2\%) & 19,520 & 39.8 & $2.06\times$ \\ \midrule
\textbf{DyTopo (Ours)} & 5 $\times$ 2.6 & \textbf{92.07 (+6.3\%)} & 9,453 & 22.3 & 1.00$\times$ \\ \bottomrule
\end{tabular}%
\end{table}

\section{Qualitative Analysis}
\label{app:qualitative}

\subsection{Topology Evolution Trace: A Case Study}
To demonstrate how DyTopo adapts to the reasoning stage, we analyze a trace from the HumanEval dataset (Problem: \texttt{is\_palindrome} and \texttt{make\_palindrome}). The system involved 4 worker agents: Researcher, Developer, Tester, and Designer.

\textbf{Round 1: Initial Exploration \& Algorithm Selection.}
\begin{itemize}
    \item \textbf{Goal:} Define initial approach.
    \item \textbf{Dynamics:} The Researcher (A) proposed KMP/Manacher algorithms. The Developer (B) drafted an initial implementation.
    \item \textbf{Topology:} A strong edge was formed from \textbf{Researcher $\to$ Developer} (Score: 0.52).
    \item \textbf{Reasoning:} The Developer's query (``I need an efficient algorithmic approach and complexity guidance'') matched the Researcher's key (``I provide candidate algorithms and complexity considerations''), routing algorithmic guidance to implementation.

\end{itemize}

\textbf{Round 2: Implementation \& Verification.}
\begin{itemize}
    \item \textbf{Goal:} Verify code against specific test cases.
    \item \textbf{Dynamics:} The focus shifted to correctness. The Tester (C) generated a comprehensive test suite.
    \item \textbf{Topology:} A critical high-confidence edge emerged: \textbf{Developer $\to$ Tester} (Score: 0.77).
    \item \textbf{Reasoning:} The Tester explicitly queried: I need the Developer's implementation to verify, which semantically aligned perfectly with the Developer's offer: I provide complete Python implementation. This routed the code directly to the Tester for validation, bypassing irrelevant agents.
\end{itemize}

\textbf{Round 3: Finalization \& Convergence.}
\begin{itemize}
    \item \textbf{Goal:} Final formatting and docstring compliance.
    \item \textbf{Dynamics:} The Manager recognized the tests passed and requested a final formatted output.
    \item \textbf{Topology:} The graph became sparse. The Manager aggregated the final outputs, and the system converged to \texttt{is\_complete=True}.
\end{itemize}

This evolution—from broad algorithmic discussion to targeted verification—confirms that DyTopo successfully aligns communication structure with the stage-dependent needs of the task.

\subsection{Examples of Semantic Key-Query Matching}
Table \ref{tab:qk_examples} presents real examples extracted from the experiment logs, illustrating how natural language descriptors are used to induce connectivity.

\begin{table}[h]
\centering
\caption{Real-world examples of Key ($s_k$) and Query ($s_q$) descriptors generated during the \texttt{is\_palindrome} task and the resulting edge formation.}
\label{tab:qk_examples}
\setlength{\tabcolsep}{4pt}
\renewcommand{\arraystretch}{1.2}
\begin{tabular}{@{}c p{5.6cm} p{5.6cm} p{4cm}@{}}
\toprule
\textbf{Round} & \textbf{Query Descriptor ($s_q$)} & \textbf{Key Descriptor ($s_k$)} & \textbf{Outcome} \\ \midrule
1 & \textbf{Agent A (Researcher):} I need the function signature for make\_palindrome implementation details. & \textbf{Agent B (Developer):} I provide the implementation using KMP failure function on the concatenated string... & \textbf{Edge Created} ($B \to A$) \newline Score: 0.52 \\ \midrule
2 & \textbf{Agent C (Tester):} I need the Developer's implementation of make\_palindrome to verify against these test cases. & \textbf{Agent B (Developer):} I provide the complete Python implementation of make\_palindrome and is\_palindrome. & \textbf{Edge Created} ($B \to C$) \newline \textbf{Score: 0.77 (High)} \\ \midrule
2 & \textbf{Agent B (Developer):} I need the Tester's report on the docstring tests and edge cases. & \textbf{Agent C (Tester):} I provide comprehensive test suite covering empty, single char, palindromes... & \textbf{Edge Created} ($C \to B$) \newline Score: 0.49 \\ \bottomrule
\end{tabular}
\end{table}

\end{document}